\documentclass[10pt,conference]{IEEEtran}
\IEEEoverridecommandlockouts
\usepackage[T1]{fontenc}
\usepackage[utf8]{inputenc}
\usepackage{amsmath,amssymb}
\usepackage{graphicx}
\usepackage{booktabs}
\usepackage{array}
\usepackage{cite}
\PassOptionsToPackage{hyphens}{url}
\usepackage{url}
\usepackage{microtype}
\usepackage[hidelinks]{hyperref}

\DeclareUrlCommand\task{\urlstyle{tt}}
\DeclareUrlCommand\code{\urlstyle{tt}}
\newif\ifsaveline \savelinetrue
\ifdefined\AddToHook
\AddToHook{para/begin}{\ifsaveline\looseness=-1 \fi}
\AddToHook{env/table/begin}{\savelinefalse}
\AddToHook{env/table*/begin}{\savelinefalse}
\fi
\newif\ifkeepcuts \keepcutsfalse
\newcommand{\cut}[1]{\ifkeepcuts #1\fi}

\begin{document}

\title{What Does Skill Lift Actually Measure?\\ A Source-Code-Level Audit and\\ Noise-Aware Reanalysis of SkillsBench\thanks{This work has been submitted to the IEEE for possible publication. Copyright may be transferred without notice, after which this version may no longer be accessible.}}

\author{\IEEEauthorblockN{Haozhe Zhang}
\IEEEauthorblockA{Trine University\\
hzhang266@my.trine.edu}
\and
\IEEEauthorblockN{Zhen Long}
\IEEEauthorblockA{National University of Singapore\\
e1351306@u.nus.edu}}

\savelinefalse
\maketitle
\savelinetrue

\begin{abstract}
Agent skills are directories of instructions, reference materials, and scripts loaded on demand by coding agents. SkillsBench reports that human-curated skills raise pass rates by 16.6 percentage points (pp) on average while lowering scores on some tasks, prompting recent work to gate or retire skills using per-task gains. We audit SkillsBench at the source-code level and reanalyze 8,769 public trials across 1,531 paired ``(configuration, task)'' cells, scoring unscored timeout trials as failures (0). Two controlled-verified grader dependencies and one source-identified operational-definition gap account for 17\% of the summed lift, averaging 68.4\,pp; excluding them reduces the mean lift from 17.6 to 15.3\,pp. Four tasks also retain placeholder skill directories, and OpenHands injects 48 built-in skills in 155 of 184 audited with-skill trajectories. Across the 83 observable tasks, overall gains are genuine (30 tasks are significantly positive at 10\% FDR) and no task shows a significant loss. Although apparent-loss cells remain negative on held-out trials ($-6.9$\,pp, 95\% CI $[-12.9,-0.7]$)---indicating that real negative effects exist---60\% of cells have zero baseline where loss is impossible, and at most three trials per condition only detect per-task losses beyond roughly 24--30\,pp, so raw loss counts cannot identify harmed tasks. Finally, baseline-conditioned gates appear to gain 1.7--2.8\,pp in offline replay, yet still gain 3.5--6.1\,pp after permuting condition labels, and yield no significant improvement on held-out trials (all upper bounds $<1$\,pp). We release our audit ledger, normalized tables, prompt patches, and reanalysis scripts.
\end{abstract}


\section{Introduction}
Coding agents such as Claude Code, Codex, Gemini CLI and OpenHands~\cite{wang2025openhands} can achieve specialisation without modifying the model weights by mounting agent skills; some directories contain a \code{SKILL.md} instruction file, as well as optional reference documents and helper scripts. At the start-up, only the short names and descriptions of all the skills are loaded into the context; then, based on relevance assessment by the agent, the entire content of the skill and related files are read~\cite{zhang2025skills}.

SkillsBench~\cite{li2026skillsbench} is the first large-scale benchmark of this mechanism. Its fourth version (v4) evaluated 87 containerised tasks across 18 model--harness configurations with and without human-curated skills, reporting a 16.6\,pp average pass-rate gain alongside wide cross-configuration variance and net drops on 13 of the 87 tasks. Building on SkillsBench, recent studies propose retiring skills that models no longer need~\cite{wang2026beyondgain}, compressing skill packages~\cite{gao2026skillreducer}, routing or retrieving skills~\cite{liu2026implicitrouting,sun2026skillrep}, measuring test coverage~\cite{tan2026skillcoverage}, and optimising skill content~\cite{ma2026skillneo}, often treating observed per-task gains and drops directly as action signals.

These studies rest on two empirical assumptions that we test directly: first, what do the largest observed gains actually measure; and second, are the apparent performance drops genuine, and can selective loading prevent them? We began with the second question by testing a gate that skips skills whenever the no-skill score is already high; although it looked effective in offline log replay, the apparent gains persisted even after permuting condition labels to erase any skill effect.

Specifically, this paper makes four contributions:
\begin{itemize}
\item \textbf{Source-code-level audit:} We cross-checked task prompts, reference solutions, graders, and skill packages in Section~\ref{sec:audit}, finding three tasks where skills supply grading rules or definitions absent from the prompt (two verified via controlled grader runs and one via source inspection).\cut{ Four tasks also retain placeholder directories, and one execution harness injects 48 built-in skills in most runs.}
\item \textbf{Noise-aware loss re-analysis:} Under within-cell permutation tests at 10\% FDR across all 83 observable tasks, 30 tasks show significant gains and none shows a significant loss (Section~\ref{sec:loss}). Although apparent-loss cells remain negative on held-out trials ($-6.9$\,pp), total negative cells fall below the zero-effect noise expectation and $\le 3$ trials per condition detect per-task losses only above roughly 24--30\,pp, so raw counts cannot identify which tasks are harmed.
\item \textbf{Evaluation trap in skill gating:} In offline replay of baseline-conditioned gates, the same noisy baseline score informs both the loading decision and the fallback outcome; label-shuffled null distributions and held-out trial scoring expose the resulting gain inflation (Section~\ref{sec:gate}).
\item \textbf{Public research assets:} We have released the normalised run and cell tables, source-code audit ledger, proposed prompt patches, and reanalysis scripts.
\end{itemize}

\section{Background and Data}
\label{sec:data}
\subsection{Task Structure}
\leavevmode\looseness=0 Each SkillsBench task consists of four components: the task prompt given to the agent (\code{task.md}), the reference solution (\code{oracle/}), the scoring script (\code{verifier/test_outputs.py}), and the skill package (\code{environment/skills/}). Each configuration runs each task under both conditions, without and with the skill package. The grader returns a score within the $[0,1]$ interval, and 96\% of the runs in our analysis scored exactly 0 or 1. According to its construction guidelines, the benchmark prohibits skills from leaking information specific to task answers, such as expected outputs or magic constants~\cite{li2026skillsbench}.

\subsection{Run Records and Paired Cells}
Two public per-run exports available in the frozen snapshot form our sample. First, a retrospective archive of early runs (\code{paper_v1_manifest.json}, generated June 15, 2026; 2,830 rows across 7 configurations of Claude Code, Codex, and Gemini CLI) was pre-filtered upstream to the earliest three non-timeout runs with finite scores and early logs per condition and task. Second, a leaderboard export dated June 8, 2026 (\code{selected_trials.csv}) adds 6,590 rows across 22 configurations, 19 on OpenHands. Throughout this paper, ``first'' and ``second'' denote these early and late run sets rather than file dates. We retain all runs on audited repository tasks; in the second export, 1,029 timeout runs without grader scores are scored as 0 (failure), matching the 385 of 387 scored paired timeouts that scored 0 (the first export filtered timeouts out upstream, and Section~\ref{sec:loss} gives the sensitivity analysis excluding unscored timeouts). A ``(configuration, task) cell'' is kept whenever both conditions are present (Table~\ref{tab:data}).

Within each cell, let $r_{0}$ and $r_{1}$ denote mean scores without and with skills, and let $\Delta=r_{1}-r_{0}$; a task's lift is the mean of $\Delta$ across its cells. Two structural constraints govern our analysis: each cell has at most three runs per condition (83 cells have only one run on at least one side), and because the second export predates the benchmark paper's final subset, partial OpenHands coverage leaves four repository tasks without paired cells. All results reflect this frozen snapshot\cut{; adding newer exports requires re-running normalisation, feature extraction, and input-hash registration}.

\begin{table}[t]
\caption{Run Record Accounting Table for the Analyzed Snapshot}
\label{tab:data}
\centering
\begin{tabular}{@{}lr@{}}
\toprule
Statistical item & Value \\
\midrule
Raw run rows of the two exports & 9,420 \\
\quad Only one condition in the cell & 574 \\
\quad Task not in the audited repository & 77 \\
Paired runs included in the analysis & 8,769 \\
\quad Timed out without a score, scored 0 & 1,013 \\
\quad Score exactly 0 / exactly 1 / partial & 5,474 / 2,914 / 381 \\
Configurations (first / second batch export) & 28 (7 / 21) \\
Tasks with at least one paired cell & 83 of 87 \\
Paired cells (first / second batch export) & 1,531 (471 / 1,060) \\
\quad Three runs under both conditions & 1,272 \\
\quad Only one run under at least one condition & 83 \\
Average score without / with skill & 26.48\% / 44.07\% \\
Cells with $\Delta>0$ / $\Delta=0$ / $\Delta<0$ & 598 / 727 / 206 \\
\bottomrule
\end{tabular}
\end{table}

\section{Source-Code-Level Audit}
\label{sec:audit}
\subsection{Grader Requirements Declared Only in Skills}
Although benchmark rules forbid skills from leaking task answers, we examine a distinct issue: grading requirements enforced by the verifier and documented in the skill but omitted from the task prompt. Cross-checking these four components yields three cases with documented evidence (Table~\ref{tab:contracts}): two verified by controlled grader runs and one by source inspection.

\leavevmode\looseness=0 \emph{Specific map projection:} In \task{earthquake-plate-calculation}, the grader accepts distances only within 0.01\,km of the reference value (about 2.6 parts per million). The accompanying skill specifies the exact projection used to compute that reference value (EPSG:4087), whereas the prompt tells agents to use GeoPandas projections without naming one.

\emph{Spreadsheet cached values:} In \task{protein-expression-analysis}, the prompt asks for Excel formulas, yet the grader opens workbooks with \code{data_only=True}, reading only cached values. Formulas written by Python libraries without auto-evaluation appear empty unless cached; the accompanying spreadsheet skill explicitly directs agents to recalculate cached values before saving.

\emph{Threshold and statistical convention:} In \task{flood-risk-analysis}, the prompt gives no operational definition of ``flood''. The accompanying skill specifies the U.S. National Weather Service (NWS) minor flood stage, defines a flood day by daily maximum reaching that threshold, and explains how to fetch station thresholds, while the reference solution fixes a threshold table and the grader asserts fixed station and day counts (even though the skill lists no per-station numbers). Unlike the first two cases, this gap rests on source-code inspection rather than a controlled grader run.

\begin{table*}[t]
\caption{The Three Retained ``Instruction--Evaluation Gap'' Cases; Evidence Levels Differ (Summarized With Equal Weight per Cell)}
\label{tab:contracts}
\centering
\begin{tabular}{@{}l r r r >{\raggedright\arraybackslash}p{5.9cm} l@{}}
\toprule
Task & Cells & \multicolumn{1}{c}{No/With skills} & Lift & Scoring-requirement gap & Evidence type \\
 & & \multicolumn{1}{c}{score (\%)} & (pp) & & \\
\midrule
\task{earthquake-plate-calculation} & 26 & 3.8 / 71.2 & $+67.3$ & Distance tolerance 0.01\,km; among three tested methods, only EPSG:4087 passes all tests & Controlled run \\
\task{protein-expression-analysis} & 18 & 9.3 / 79.6 & $+70.4$ & Grader reads cached cell values; unevaluated formulas read as empty & Controlled run \\
\task{flood-risk-analysis} & 22 & 1.5 / 69.7 & $+68.2$ & No specified flood-stage source, daily-maximum statistic, or threshold comparison convention & Source-code review \\
\midrule
Total for three tasks & 66 & 4.5 / 73.0 & $+68.4$ & & \\
\bottomrule
\end{tabular}
\end{table*}

\textbf{Controlled validation:} For the first two tasks, we used the official grader on candidate outputs that differed only by an undeclared requirement. Spherical great-circle distance (3,552.98km), Web Mercator distance (4,152.56km), and EPSG:4087 distance (3,878.27km) of the earthquake catalog data all point to the same seismic event; however, only EPSG:4087 passed all eight assertions, and the other two failed the distance check and scored 0. We tested two versions of the same protein-expression workbook. Both contained identical formulas in all 143 formula cells. The version without cached values passed three of the six tests, and the version with cached values passed all six.

\textbf{Interpretation of results:} Together, these three tasks represent 4\% of all paired cells (66/1,531), contain no loss cells, and descriptively account for 17\% of summed lift; removing them lowers the cell-weighted mean lift from 17.6 to 15.3\,pp. These summaries weight paired cells equally and represent neither causal effects nor task-weighted means. In both controlled cases, varying only the undeclared requirement flipped the grader outcome, though these tests do not isolate how much of each task's lift comes from that requirement rather than broader procedural guidance. Nor do we suggest these skills were tailored to the grader: for \task{protein-expression-analysis}, recalculating cached values is legitimate procedural knowledge, and the flaw lies in a grader reading only cached values and penalizing workbooks that Excel would evaluate upon opening. Still, part of the measured gain reflects undocumented grading contracts. Separating procedural help from rule disclosure requires running agents with prompts that state those contracts explicitly; we release prompt patches for all three tasks to enable that control. Finally, this audit is not exhaustive: the next two highest-lift tasks---\task{llm-prefix-cache-replay} (0 to 78.1\%) and \task{dapt-intrusion-detection} (7.4 to 75.9\%)---remain natural follow-ups.

\leavevmode\looseness=0 \textbf{Unconfirmed cases:} Further investigation eliminated the two previous candidate tasks. The grading script of \task{weighted-gdp-calc} calls \code{ssconvert} before result verification and recovers computed CSV values even without Excel caches; thus, the presence of \code{data_only=True} did not require caching for the agent. Assets-under-management tolerance in \task{sec-financial-report} and the helper script for rounding to two decimal places did not constitute an undeclared unit or rounding trap. Neither task has controlled evidence of an ``instruction-evaluation gap''.

\subsection{Skills Pointing to Different Algorithms}
Skill-package guides the agent to an algorithmic path that avoids the official reference solution in the three tasks.

For \task{exam-block-sequencing}, the skill gives a three-index mixed-integer programming (MIP) model. Given the actual input scale of the task (24 blocks, 22 windows, and 11,130 distinct block triples), only the assignment variables and window variables need 245,436 binary variables and 979,440 linearization constraints. Conversely, as mentioned in the reference solution, instead of employing this large MIP, simulated annealing over block permutations combined with exact dynamic programming for slot assignment was utilised.

Skill passes a custom Python function as the distance measure for the clustering process of \task{mars-clouds-clustering}. Over the seven timed benchmarks in this task's dataset, a single round with the Python callback has a median time of 1.67 seconds and that of the reference solution's vectorized distance computation is 0.027 seconds; over the task's 847-point parameter grid, distance evaluation alone extrapolates to 23.6 minutes compared with 23 seconds.

For \task{r2r-mpc-control}, some of the skill package teaches finite-horizon and receding-horizon model predictive control (MPC), and the reference solution solves a single infinite-horizon linear-quadratic regulator problem. However, even the MPC tuning skill of this package uses a continuous-time algebraic Riccati solver; thus, it is neither fully finite-horizon nor in direct opposition to the reference method.

These three cases illustrate alternative solution strategies rather than defective skills, as the reference solution is only one valid path. In the second export, paired with-skill and no-skill timeout counts are comparable on \task{exam-block-sequencing} (25/46 vs.\ 23/46) and \task{mars-clouds-clustering} (16/41 vs.\ 13/39), showing no timeout penalty from mounting the skills, though we did not inspect trajectories to verify which algorithm agents executed. As shown in Section~\ref{sec:loss}, none of the three tasks shows a statistically significant loss in pooled tests, and two show positive lift in the first export.\cut{ We retain them as concrete hypotheses for future trajectory-level analysis, consistent with qualitative failure modes noted in SkillsBench~\cite{li2026skillsbench}.}

\subsection{Placeholder Directories and Non-Skill Files}
Each of four tasks contains a directory named after a job title, such as \task{senior-data-engineer}, alongside its task-specific skills. Together, these four directories contain 53 files totaling approximately 584 KiB. Nine files across two of the directories contain literal code-generator placeholders such as ``TODO: Implement actual functionality''. Two more tasks mount a \code{licenses/} subdirectory for the third-party software license texts.

We found no evidence that these residual files harm task performance: the four placeholder tasks have lifts between $0.0$ and $+12.1$\,pp, the two license tasks improve by $+17.9$ and $+10.0$\,pp, and the baseline-stratified comparison in Section~\ref{sec:loss} detects no significant difference in loss rates. This is consistent with progressive disclosure---full text of unopened files never enters the model context (even if brief directory descriptions do), so raw directory size is not a good proxy for context load. Nevertheless, these residual files do pass upstream validation: although the repository's continuous integration checks skill frontmatter, existing checks do not block code-generator placeholder files.

\subsection{Harness That Injects Built-in Skills}
Upstream maintainers extracted the first-turn system prompts of 190 OpenHands trajectories during the review\cut{ and found that there is a considerable built-in overhead in the skill directories of these prompts}. Out of the 184 with-skill trials, 155 prompts listed 48 built-in OpenHands skills before 1 to 7 task-specific skills (reconstructed directory text is 15,449 characters long and does not include token count), and task-specific skills accounted for only 2\%-13\% of the directory entries. The same 48 built-in skills appeared in 4 of the 6 no-skill trials. Among the 184 with-skill trials, 155 listed these built-in skills and the remaining 29 listed only task-specific skills.

Table~\ref{tab:harness} shows the pass rates of the three models on the leaderboard for both native CLI and OpenHands, with a mean lift of 20.9\,pp for native CLI and 15.3\,pp for OpenHands. Three model pairs cannot isolate built-in catalog dilution as the cause of this gap, since one model already has a higher no-skill baseline under OpenHands. Rather, because even ``no-skill'' OpenHands trials still expose built-in skills, skill lift is not a strict like-for-like metric across harnesses.

\begin{table}[t]
\caption{Leaderboard No-Skill/With-Skill Pass Rates (\%) and Lifts (pp)}
\label{tab:harness}
\centering
\begin{tabular}{@{}lcc@{}}
\toprule
Model & Native CLI & OpenHands \\
\midrule
GPT-5.5 & 46.8 / 66.5 ($+19.7$) & 51.5 / 67.3 ($+15.8$) \\
Claude Opus 4.7 & 43.0 / 61.2 ($+18.2$) & 42.1 / 53.1 ($+11.1$) \\
Gemini 3.1 Pro & 36.0 / 60.8 ($+24.8$) & 33.8 / 52.8 ($+19.0$) \\
\bottomrule
\end{tabular}\\[2pt]
\parbox{\columnwidth}{\footnotesize Lifts are taken from the official columns of unrounded raw ratios on the leaderboard.}
\end{table}

\subsection{Upstream Leak Audit Records}
Upstream repository also contains the maintainer audit records of 2,333 trials, and 58 of them are marked as anomalous skill loading or no-skill leakage, and 7 as suspected score gaming or grader leakage (not all of which are confirmed grader access). Each of the flagged records has a trial ID and path; however, 33 of them involve two models not in our snapshot, and matching the rest of the columns in our export table yielded no results. As identifier non-matching does not guarantee that the two datasets are entirely separate, we have not reported a contamination rate for our sample and have only mentioned that cross-condition isolation requires explicit confirmation.

\section{Are the Losses Real?}
\label{sec:loss}
In our snapshot, 13.5\% of cells (206/1,531) satisfy $\Delta<0$, and 13 tasks show net negative lift. How much of these apparent regressions exceeds random trial noise?

\subsection{When Losses Can Occur}
The sign of $\Delta$ is strictly constrained by the cell's no-skill baseline $r_{0}$ (Table~\ref{tab:sign}). In the 921 cells where all no-skill trials fail ($r_{0}=0$), a loss is arithmetically impossible. Conversely, in the 210 cells where all no-skill trials pass ($r_{0}=1$), a positive lift is impossible, and a single failure among three with-skill trials produces a 33\,pp apparent drop. Thus losses can occur only in the 610 cells with positive baseline---within this subset, 34\% (206/610) show $\Delta<0$.

Comparing the two exports underscores how multiple factors shift together. While the first export runs exclusively on native CLIs, 18 of the 21 paired configurations in the second export use OpenHands, and only the second export retains timeouts. Once unscored timeouts are scored as failures, the second and first exports share nearly identical no-skill baselines (26.4\% vs.\ 26.7\%) and all-pass cell rates (13.8\% vs.\ 13.6\%), yet the second export exhibits lower mean lift (15.6 vs.\ 22.0\,pp) and more loss cells (15.0\% vs.\ 10.0\%). Because models, harnesses, and upstream timeout filtering change together, these differences cannot be attributed to a single factor.

\begin{table}[t]
\caption{Sign Distribution of $\Delta$ Stratified by Cell No-Skill Score $r_{0}$}
\label{tab:sign}
\centering
\begin{tabular}{@{}lrrrr@{}}
\toprule
No-skill score $r_{0}$ & Total cells & $\Delta<0$ & $\Delta=0$ & $\Delta>0$ \\
\midrule
Exactly 0 & 921 & 0 & 519 & 402 \\
Between 0 and 1 & 400 & 135 & 69 & 196 \\
Exactly 1 & 210 & 71 & 139 & 0 \\
\midrule
All cells & 1,531 & 206 & 727 & 598 \\
\bottomrule
\end{tabular}
\end{table}

\subsection{Task-Level Statistical Tests}
For each observable task, we test the strict null hypothesis that skills have no effect in any of its cells. Within each cell, we randomly permute condition labels across trials while holding the number of trials per condition fixed, then recompute task lift. Using 100,000 permutations with add-one correction, we compute two-sided Monte Carlo $p$-values for all 83 tasks, and control the false discovery rate (FDR) at 10\% via the Benjamini--Hochberg method~\cite{benjamini1995fdr}.

\begin{figure}[t]
\centering
\includegraphics[width=\columnwidth]{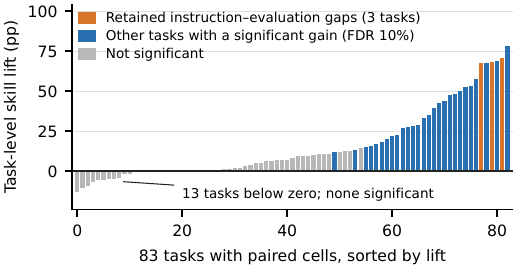}
\caption{Distribution of lift across the 83 tasks with paired cells. Significance comes from two-sided within-cell label permutation tests controlling the false discovery rate at 10\% across all 83 tasks.}
\label{fig:tasks}
\end{figure}

A total of 30 tasks reach significance, and without exception all are positive lifts (Fig.~\ref{fig:tasks}). No negative-lift task comes close to significance: even before multiple-testing correction, the smallest two-sided $p$-value among all 13 negative tasks is 0.14 (one-sided 0.071). For example, the most negative task overall, \task{radar-vital-signs} (mean $-13.3$\,pp across 15 cells), contains only 5 loss cells, 2 gain cells, and 8 ties ($p=0.21$). Of course, non-significance does not prove a task is unharmed; with at most three trials per condition, task-level tests have low statistical power, which Section~\ref{sec:discussion} quantifies.

\leavevmode\looseness=0 Batch-specific estimates vary markedly across exports. Among the 8 overall negative-lift tasks observed in both exports, 4 flip sign between them; for instance, \task{r2r-mpc-control} moves from $+14.3$\,pp (7 cells) in the first export to $-9.7$\,pp (12 cells) in the second, despite identical skill files across both audited commits. Because the exports differ in models, harnesses, and timeout filtering, such reversals capture cross-setting instability as well as trial noise. Even within the second export alone, the most negative task---\task{multilingual-video-dubbing} ($-18.3$\,pp across 10 cells)---reaches only an uncorrected two-sided $p=0.15$.

\subsection{How Many Negative Cells Would Pure Noise Produce?}
The same permutation null distribution also indicates how many negative cells arise by chance under zero skill effect. Exact enumeration of within-cell label assignments yields 367.9 expected loss cells (central 95\% range 341--395) and 365.0 expected gain cells, versus 206 loss and 598 gain cells in the data---significantly fewer losses than under the global zero-effect benchmark. This inequality does not imply all skills are benign: a mix of many helpful skills and a few harmful ones would still fall below 367.9. Moreover, permuting labels inside strongly beneficial cells (e.g., $0/3$ vs.\ $3/3$) yields apparent losses half the time, inflating the null expectation and giving this global count low power to detect a small subset of genuinely harmful skills.

\subsection{Held-Out Trial Test}
To separate true effects from selection noise, we examine the 1,448 cells with at least two trials per condition. In each random split, we hold out one trial per condition, use the remaining decision trials to compute $r_{0}$, $r_{1}$, and $\Delta$, then test how the selected set of cells performs on its held-out trial pair. Point estimates are the average of 2,000 random divisions, the 95\% intervals are obtained by task-level resampling, and repeated divisions use the same underlying trial records without being independent replications.

Cells selected by high decision baseline ($r_{0}\ge0.8$, about 271 per split) have mean $\Delta=-20.8$\,pp on decision trials, but rebound to $-0.1$\,pp on held-out trial pairs (95\% interval $[-4.9,+4.7]$). This is textbook regression to the mean~\cite{barnett2005regression}: cells enter the $r_{0}\ge0.8$ group partly because their initial no-skill trials happened to succeed.

Cells selected by apparent loss on decision trials ($\Delta<0$, about 163 per split) average $-55.6$\,pp on decision trials, but $-6.9$\,pp on held-out trial pairs (95\% interval $[-12.9,-0.7]$), with 24\% showing losses and 19\% gains. As the mean of the hold-out set is still negative, apparent-loss cells are not truly random noise on average. Even so, the ratio $-6.9$ to $-55.6$\,pp should not be read directly as the fraction of real harm: the set contains different cells that exhibit different kinds of effects, each hold-out cell is based on one experiment on one side, and the observed data cannot tell apart the harm due to the skill itself from the interference between experiments in the same condition.

\subsection{Structural Features and Loss Rates}
Furthermore, we evaluate if having any of three characteristics of an audited skill package (different route of algorithm from reference solution, 3 tasks; containing placeholder directories, 4 tasks; containing license directories, 2 tasks) correlates with increased probability of having loss. Overall, these 9 flagged tasks show a higher loss rate than the other 74 tasks (20.6\% vs.\ 12.6\%), but they also have a much higher no-skill baseline (37.5\% vs.\ 25.2\%), so there is more room for loss to happen there: loss is only possible where the baseline is above zero. Restricting to positive-baseline cells ($r_{0}>0$), the gap narrows to 35.6\% (32/90) vs.~33.5\% (174/520), with a baseline-stratified Mantel--Haenszel odds ratio of 1.10 and a task-level bootstrap rate difference of $+2.1$\,pp (95\% interval $[-9.8,+16.6]$). Moreover, 11 of the 13 negative-lift tasks have none of these three structural features.

\subsection{Timeout Analysis}
In the second export, which records timeout status, paired with-skill trials time out at 23.6\% (720/3,049) versus 24.1\% (714/2,960) for no-skill trials, a ratio of 0.98 (95\% task-bootstrap interval 0.88--1.09); thus, mounting skills does not measurably alter the overall timeout rate. Restricted to cells selected by $\Delta<0$, this ratio rises to 3.6 (34.5\% vs.\ 9.7\%), illustrating outcome-selection bias under a unified denominator. Excluding the 1,029 unscored timeout rows (1,013 in paired cells) instead of scoring them as 0 lowers apparent paired timeout rates to 9.0\% and 7.1\% (a spurious ratio of 1.26), shrinks loss cells from 206 to 169, lifts the second-export baseline from 26.4\% to 31.7\%, and shifts 20 tasks' lifts by over 5\,pp (7 by over 10\,pp), though all qualitative conclusions of Sections~\ref{sec:loss} and~\ref{sec:gate} hold under both conventions.

\section{Can a Gate Conditioned on Baseline Help?}
\label{sec:gate}
Because apparent losses cluster in high-baseline cells, a natural response is selective loading---activating skills only when predicted to help. We evaluate such gates under three protocols that yield starkly different conclusions.

\subsection{Gating Strategies}
\emph{Baseline cutoff:} load skills only when $r_{0}<0.8$. This demonstrates the most straightforward interpretation of ``retire skills if the bare model can perform adequately enough''; given that Wang et al.~\cite{wang2026beyondgain} use a threshold of 0.8 for scores obtained from a separate probe dataset, with the addition of a cross-model validity filter, our threshold rule is merely a rough approximation thereof.

\emph{Random-forest gate:} two random forests~\cite{breiman2001rf} use 17 features---$r_{0}$, an OpenHands harness indicator, and 15 static descriptors covering skill-package size, file composition, and lexical overlap with the prompt---to predict expected gain $\Delta$ and loss probability $P(\Delta<0)$, respectively. Each forest contains 25 trees with maximum depth 5, producing out-of-fold predictions via 10-fold cross-validation across cells. The max-gain version loads skills when $\hat{\Delta}>0.02$, while the risk-averse version loads when both $\hat{\Delta}>0$ and $\hat{P}(\Delta<0)<0.25$. The baseline-only forest uses only the first two features, with 20 trees of maximum depth 4.

\emph{Oracle-selector for Best-of-two:} directly selects the condition with the highest observed score.

\subsection{Three Evaluation Protocols}
\emph{Replay.} Standard offline log replay evaluates on the same recorded trials: when the policy loads skills, score $r_{1}$; when it declines, score $r_{0}$.

\emph{Permuted labels.} Within each cell, we randomly shuffle condition labels across trials, recompute $r_{0}$, $r_{1}$, $\Delta$, and loss indicators from the shuffled trials, refit the random forests, and perform offline replay. Given that shuffling inherently eliminates any skill effect, any positive effect or loss removal rate must be artificially created by the assessment process. Forest policies average 60 permutations; model-free rules average 2,000 permutations.

\emph{Held-out trial.} Restricting to cells with at least two trials per condition leaves 1,448 eligible cells and excludes 83. At each split, we randomly set aside one trial per condition solely for scoring, recalculate baseline features and training labels only from the rest of the decision trials, fit out-of-fold forests on these decision trials, and finally evaluate the load/skip decisions completely on the held-out trial pair. Gate thresholds and forest hyperparameters are set according to the configurations chosen in the previous full-data exploration and are not re-tuned. For each cell, we calculate the mean of held-out gains over random splits (60 for forests, 2,000 for other rules) and resample entire tasks 10,000 times to build the 95\% confidence intervals; due to the lack of internal refitting in bootstrap resampling and shared trial pools for splits, these intervals reflect task-sampling variation under a given fitted policy rather than independent replications.

\begin{table*}[t]
\caption{Gains (pp) Relative to ``Always Load Skills'' Under Three Evaluations}
\label{tab:gate}
\centering
\setlength{\tabcolsep}{5pt}
\begin{tabular}{@{}l r r r r r r@{}}
\toprule
 & \multicolumn{2}{c}{Replay} & \multicolumn{2}{c}{Permuted-label} & Decision-set & Held-out trial \\
\cmidrule(lr){2-3}\cmidrule(lr){4-5}
Strategy & Gain (95\% CI) & Removal rate & Gain & Removal & replay gain & gain (95\% CI) \\
\midrule
Baseline cutoff & +2.65 (+1.85, +3.53) & 37\% & +3.5 & 30\% & +3.9 & +0.02 ($-$0.85, +0.90) \\
RF gate, max gain & +2.79 (+1.90, +3.73) & 55\% & +6.1 & 79\% & +4.2 & $-$0.13 ($-$1.04, +0.79) \\
RF gate, risk averse & +1.72 (+0.47, +2.96) & 91\% & +5.5 & 93\% & +3.6 & $-$0.89 ($-$1.98, +0.22) \\
RF gate, baseline only & +0.67 ($-$0.87, +2.15) & 98\% & +5.3 & 96\% & +2.9 & $-$1.43 ($-$2.68, $-$0.17) \\
Best of two & +5.46 (+4.26, +6.67) & 100\% & +9.5 & 100\% & +6.3 & $-$1.65 ($-$2.99, $-$0.29) \\
\bottomrule
\end{tabular}\\[2pt]
\parbox{0.8\textwidth}{\footnotesize ``Removal'' refers to the proportion of cells with $\Delta<0$ where the strategy does not load skills. The permuted-label column is the mean over permutations; intervals are 95\% task bootstrap intervals.}
\end{table*}

\begin{figure}[t]
\centering
\includegraphics[width=0.82\columnwidth]{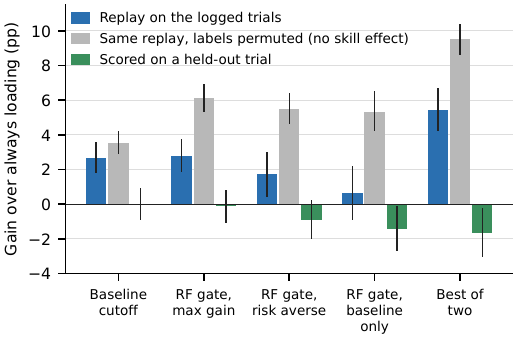}
\caption{Comparison of the three evaluation approaches outlined in Table~\ref{tab:gate}. The error bars for log replay and holdout trials represent 95\% task bootstrapped intervals, while those for permuted-label replay cover the 95\% central range of permutation results.}
\label{fig:gate}
\end{figure}

\subsection{Evaluation Results}
Table~\ref{tab:gate} and Fig.~\ref{fig:gate} compare the three evaluation methods. In standard log replay, the baseline cutoff and both 17-feature forests beat always loading skills by 1.7--2.8\,pp, and the risk-averse forest skips 91\% of apparent-loss cells. Yet on label-shuffled data with no real skill effect, those same three policies gain 3.5--6.1\,pp and skip 93\% of shuffled ``loss'' cells. On held-out trials, no strategy achieves a significant gain: confidence intervals for the cutoff and both full-feature forests cross zero with upper bounds below 1\,pp, well short of the 1.7--2.8\,pp replay claims\cut{ (though wide intervals do not prove the true effect is exactly zero)}. Worse, choosing the decision-trial winner (Best of two) loses 1.65\,pp (95\% CI $[-2.99,-0.29]$) out of sample, and the baseline-only forest likewise incurs a significant held-out loss ($-1.43$\,pp, $[-2.68,-0.17]$), by skipping cells that tie in decision trials yet benefit on held-out pairs.

The mechanism behind this illusion stems directly from $r_{0}$ entering three places simultaneously. Estimated from at most three trials, $r_{0}$ serves as a gate input feature, forms the negative half of the training target $\Delta=r_{1}-r_{0}$, and directly provides the fallback score when the gate skips loading. As a result, any cell whose no-skill trials happen to succeed is labeled ``high risk'', and those lucky passes are immediately credited to the policy. A simple calculation shows the magnitude: suppose a cell has true pass probability 0.7 both with and without skills (i.e., skills have no effect). With three no-skill trials, the probability all three pass is $0.7\times0.7\times0.7=0.34$; whenever this happens, the cutoff rule skips skills and records a score of 1.0 (while always-loading has expected score 0.7), creating about $+10$\,pp of gain on that cell from pure noise. In statistical terms, offline replay selects on a noisy baseline measurement and evaluates on the exact same random realization~\cite{cawley2010overfitting}, causing target leakage through a shared term~\cite{kapoor2022leakage}. Cross-validation across cells cannot remove this bias because the coupling occurs within each individual cell.

\subsection{What the Detector Actually Detects}
At first glance, the risk forest's ROC-AUC of 0.884 for predicting $\Delta<0$ seems to show strong ability to identify harmful skill packages. Table~\ref{tab:auc} decomposes this number and reveals a different picture. Checking only the binary indicator $r_{0}>0$ already yields AUC 0.848, because it directly excludes the 60\% of zero-baseline cells where loss is arithmetically impossible. Within the $r_{0}>0$ subset where losses can actually occur, the 17-feature forest reaches 0.62 under cell-level cross-validation but drops to 0.55 when holding out entire tasks. As the 15 static descriptors are the same for all cells in the same task, cell-level folds have used them as an implicit ``task ID'' for the forest. With folds grouped by task, the 17-feature forest trailed the baseline-and-harness forest by $0.007\pm0.005$ AUC across all cells and by $0.022\pm0.017$ among cells with $r_{0}>0$ (mean $\pm$ standard deviation of paired differences across eight random seeds), confirming that static package descriptors add no cross-task signal once implicit task identity is removed. The forests also differed in size and depth (25 depth-5 trees versus 20 depth-4 trees), so the comparison cannot isolate the contribution of the additional features. Tellingly, the full-feature forest still achieves AUC 0.858 on label-shuffled data; when predicting the sign of $\Delta$ on held-out trials, it scores 0.71, only slightly ahead of the 0.69 obtained by the decision-trial indicator $r_{0}>0$ alone.

\begin{table}[t]
\caption{AUC for Predicting Loss ($\Delta<0$)}
\label{tab:auc}
\centering
\begin{tabular}{@{}lcc@{}}
\toprule
Predictor & All cells & Cells with $r_{0}>0$ \\
\midrule
Indicator $r_{0}>0$ & 0.848 & 0.500 \\
Numeric $r_{0}$ & 0.850 & 0.506 \\
\multicolumn{3}{@{}l}{\textbf{Folds over cells}} \\
Random forest (baseline + harness) & $0.870\pm0.001$ & $0.574\pm0.004$ \\
Random forest (all 17 features) & $0.884\pm0.003$ & $0.618\pm0.011$ \\
\multicolumn{3}{@{}l}{\textbf{Folds over tasks}} \\
Random forest (baseline + harness) & $0.869\pm0.003$ & $0.570\pm0.009$ \\
Random forest (all 17 features) & $0.862\pm0.005$ & $0.548\pm0.016$ \\
\bottomrule
\end{tabular}\\[2pt]
\parbox{\columnwidth}{\footnotesize Forest rows are mean $\pm$ standard deviation across 8 random seeds; standard deviation reflects fitting-process fluctuation, not sampling uncertainty.}
\end{table}

\section{Discussion}
\label{sec:discussion}
\textbf{What does skill lift actually measure.} Three factors combine in aggregate skill lift here: genuine procedural guidance, undeclared grading conventions supplied only in skills, and cross-harness shifts in ``no-skill'' baselines. Only the first reflects what a skill benchmark aims to isolate. The three gap tasks contribute roughly 17\% of summed lift, though quantifying the exact fraction due to missing prompt rules requires re-running agents with patched prompts.

\textbf{Recommendations for benchmark maintainers.} (1) List all verifier-enforced rules in the task prompt or explicitly mark skill-dependent conventions; automated diffing of grader constants, API calls, and \code{task.md} can flag candidates, but execution-path review and controlled runs remain essential. (2) Extend skill CI checks beyond frontmatter to catch code-generator placeholders and non-skill directories. (3) Save exact skill directories shown in each run and keep non-task skills fixed across control groups. (4) Report task-level uncertainty intervals and size trials adequately. Under independent binary trials at a moderate pass rate across 18 configurations, three trials per condition yield a task-lift standard error near 10\,pp, requiring a true loss exceeding roughly 24\,pp for 80\% power in an uncorrected one-sided 5\% test (or roughly 28--39\,pp under two-sided 10\% FDR across 83 tasks); raising this to ten trials per condition lowers the uncorrected threshold to about 13\,pp.\cut{ These calculations serve only for sample-size planning rather than exact guarantees.}

\textbf{Recommendations for methods that act on gains.} Any decisions based solely on cell-level differences derived from just three experiments will not be reliable enough to retire a skill or gate it. To evaluate a selective-loading policy, score its decisions on trials withheld from fitting and decision-making within each split. Repeat the same analysis after shuffling condition labels within each cell to check for apparent gains under the null. To assess generalization to unseen tasks, assign all configuration--task cells from the same task to the same fold. Likewise, loss-detection AUC should be reported within the $r_{0}>0$ subset where losses can actually occur.

\section{Threats to Validity}
\textbf{Snapshot differences.} Our frozen snapshot differs from the benchmark paper's final subset: it covers more configurations (some on partial task subsets), and 11\% of raw rows lack finite grader scores (all second-export timeouts). Upstream filtering also differs across exports: the retrospective first archive retains only non-timeout runs with early logs, whereas the second export preserves timeout runs. In our primary analysis, we score the 1,029 unscored timeout trials on audited tasks as 0 because exceeding the time limit fails to solve the task; excluding them instead alters task rankings and loss counts (Section~\ref{sec:loss}) without altering any main statistical conclusion. Consequently, exact counts can differ on the final dataset, and cross-export differences should not be attributed solely to time or model capability. Although there are only about three experiments per cell per condition for sampling noise, the main cause of this noise is the same in the final data; however, the degree of noise varies according to pass rates and within-cell correlation.

\textbf{Held-out scoring variance.} Held-out evaluation scores decisions with only a single-trial pair per cell and excludes 83 low-trial-count cells.\cut{ Due to the noise of a single split, we average over repeated splits (60 for forests, 2,000 for other rules) and bootstrap-resample across tasks; however, repeated splits reuse the same trials instead of providing new replications, and bootstrap intervals do not account for variation due to refitting forests on new task samples.} Finally, intervals-crossing-zero only bound the strategy's gain and do not prove that it is exactly zero, as we did not specify an equivalence margin.

\textbf{Within-cell trial dependence.} Within-cell label permutation assumes trials under a given cell are exchangeable under the null. If there is correlated trial interference---for example, transient API or container failures affecting multiple trials in one condition---exchangeability is violated, and $p$-values or intervals may be skewed in either direction.

\textbf{Audit coverage.} The scope of our source audit is limited; two retained gap tasks are supported by controlled grader experiments, \task{flood-risk-analysis} is based on source-code review, and the earlier candidate tasks (\task{weighted-gdp-calc} and \task{sec-financial-report}) have not been verified.\cut{ As we have not conducted new-agent trajectory experiments, we cannot determine the amount of lift that would be retained when prompts specify grading rules.}\cut{ OpenHands directory and leak counts are also based on maintainer-reviewed exports instead of re-extraction from raw trajectories.}

\textbf{Retrospective policy setting.} After observing the run results, skill-package structural markers were coded; then, in an earlier analysis phase when all trials (including later held-out trial pairs) were visible, forest hyperparameters and gate thresholds were selected and fixed. Therefore, our hold-out protocol is a retrospective robustness test of a pre-defined fixed policy, not a prospective experiment on untouched new data; we conducted no secondary tuning, and thus do not rule out the possibility of training an effective gate with an independent probe set.

\section{Related Work}
Li et al.~\cite{li2026skillsbench} introduced the paired with/without-skill evaluation benchmark for coding agents in SkillsBench and recorded cross-harness variation and task-level declines; this paper provides a source-level grading audit and a statistical noise analysis of these declines. Wang et al.~\cite{wang2026beyondgain}, the closest prior study, show that raw gain scores confound skill effect with baseline capability and propose retiring skills via dedicated bare-model probes; our held-out evaluation reinforces their warning and shows that baseline coupling also distorts offline replay. Concurrent work has also studied skill compression~\cite{gao2026skillreducer}, skill routing and representation~\cite{liu2026implicitrouting,sun2026skillrep}, test-sufficiency criteria~\cite{tan2026skillcoverage}, evolutionary skill optimisation~\cite{ma2026skillneo} and safety oversight instructions~\cite{sanko2026oversight}.

More generally, in code generation~\cite{liu2023evalplus}, repository issue fixing~\cite{jimenez2024swebench,chowdhury2024swebenchverified}, question answering~\cite{gema2025mmluredux}, and standard classification datasets~\cite{northcutt2021pervasive}, post hoc source audits have repeatedly reshaped how benchmarks are interpreted. Recent agent-benchmark checklists stress both task validity and outcome validity~\cite{kapoor2024agents,zhu2025abc}, and hidden grader rules directly compromise task validity. At the statistical level, our analysis applies three established mechanisms to skill evaluation: regression to the mean~\cite{barnett2005regression}, selection bias in model evaluation~\cite{cawley2010overfitting}, and target leakage through shared variables~\cite{kapoor2022leakage}.

\section{Conclusion}
Several of the largest skill gains in this SkillsBench snapshot occur on tasks with instruction--evaluation gaps, as shown by two controlled grader tests and one source-level definition gap. Conversely, although apparent-loss cells remain negative on average in held-out trials, negative cells occur less often overall than expected from pure noise under zero effect, and no task shows a significant loss after multiple-testing correction. Likewise, baseline-conditioned gates appear promising in naive replay yet show no significant held-out gain, with two policies (the baseline-only forest at $-1.43$\,pp and Best of two at $-1.65$\,pp) incurring significant held-out losses. These findings apply strictly to the fixed policies and frozen snapshot analysed here, neither proving skills universally harmless nor ruling out principled gating. They do support concrete methodological practices: benchmarks should declare grader-enforced rules explicitly in prompts and report uncertainty intervals, and gain-driven methods should validate on held-out trials against label-shuffled nulls.

\section*{Data Availability}
The repository at \url{https://github.com/rrrrr4788/skillsbench-lift-reanalysis} provides normalised trial and cell tables with exclusion reasons, the source audit ledger, prompt patches for Table~\ref{tab:contracts}, controlled grader scripts, static checkers for placeholder and non-skill directories, and the full reanalysis pipeline.\cut{ All tables and figures are reproducible from these materials, with output files recording random seeds, package versions, input hashes, and per-split held-out trial IDs.}

\section*{Acknowledgment}
The authors used AI assistants (Claude and Gemini) to assist with running verification scripts and polishing English grammar. All study design, source-code audit judgments, statistical analyses, and conclusions are the work of the authors, who independently verified every reported result and take full responsibility for the manuscript.

\bibliographystyle{IEEEtran}
\bibliography{IEEEabrv,refs}

\end{document}